\documentclass[letterpaper, 10 pt, conference]{ieeeconf}
\usepackage{cite} 
\usepackage{multirow}
\usepackage[utf8]{inputenc}
\usepackage[misc]{ifsym}
\usepackage{graphicx}
\usepackage{tikz}
\usepackage{pifont}
\usepackage[misc]{ifsym}
\usepackage{tikz}
\usepackage{bbding}
\usepackage{pdfpages}
\usepackage{amsmath}
\usepackage[utf8]{inputenc}
\usetikzlibrary{arrows, shapes, positioning, shadows, trees, calc, decorations.markings}

\usepackage{amssymb}
\usepackage{hyperref}
\usepackage{booktabs}
\usepackage{algorithm}
\usepackage{algpseudocode}
\usepackage{balance}

\usepackage{color, colortbl}
\definecolor{LightCyan}{rgb}{0.88,1,1}

\hypersetup{
colorlinks=true,
linkcolor=black
}

\makeatletter
\newcommand{\printfnsymbol}[1]{%
  \textsuperscript{\@fnsymbol{#1}}%
}

\makeatother

\IEEEoverridecommandlockouts                

\begin{document}

\title{\LARGE \bf
BiliVLA: Scene-Aware Vision-Language-Action Model with Reinforcement Learning for Autonomous Biliary Endoscopic Navigation\\
}
\author{
Jinsong Lin$^{1,*}$, Chi Kit Ng$^{1,*}$, Zhiyong Xiong$^{3,*}$, Zikang Pan$^{1}$, Yihan Hu$^{4}$, Tamima Tabassum$^{1}$,\\
Ziyi Hao$^{1}$, Eddie Cheung$^{5}$, Jiewen Lai$^{1}$, Huxin Gao$^{1}$, and Hongliang Ren$^{1,2,\dagger}$%
\thanks{This work was supported in part by the Ministry of Science and Technology (MOST) of China Key Project (2025YFE0122500 and 2024YFE0216200); the Innovation and Technology Fund (ITF) of Hong Kong SAR (ITF MHP/185/24); the Hong Kong Research Grants Council (RGC) General Research Fund (GRF 14216022, 14204524, 14203323, and 14206125); the RGC Research Impact Fund (RIF R4020-22); and the RGC Strategic Topics Grant (STG 1/M-405/25-N); CUHK Direct Grant for Research 2024/2025 (4055262); Young Talent Support Project of Guangzhou Association for Science and Technology (QT-2025-046).}
\thanks{$^{*}$These authors contributed equally.}%
\thanks{$^{\dagger}$Corresponding author: {\tt\small hlren@ee.cuhk.edu.hk}}%
\thanks{$^{1}$The Chinese University of Hong Kong, Hong Kong SAR, China.}%
\thanks{$^{2}$Shenzhen Loop Area Institute, China.}%
\thanks{$^{3}$The Third Affiliated Hospital of Sun Yat-sen University, Guangzhou, China.}%
\thanks{$^{4}$University of Cambridge, Cambridge, United Kingdom.}%
\thanks{$^{5}$University of California, Davis, Davis, CA, United States.}%
}


\maketitle

\begin{abstract}
Endoscopic retrograde cholangiopancreatography (ERCP) demands precise endoscopic navigation and stable biliary cannulation within a narrow monocular field characterized by specular reflections, partial occlusions, and frequent tissue contact. Although recent robotic systems and vision-based assistance techniques improve operator ergonomics and provide perceptual cues, their performance degrades under pronounced anatomical variability and safety-critical visual artifacts, which hinders reliable autonomy in cannulation-grade procedures. 
Here, we present BiliVLA, a scene-aware Vision-Language-Action (VLA) framework that formulates biliary endoscopic navigation as an instruction-conditioned visuomotor learning problem. Given an endoscopic observation and a stage-specific language instruction, BiliVLA jointly predicts the target category, a grounded bounding box, and a discrete three-degree-of-freedom (3-DoF) motor command for a continuum endoscope. The proposed framework incorporates scene-aware supervision to improve semantic target consistency and safety-aware recovery supervision to induce conservative retreat behaviors under luminal wall contact. A key component of BiliVLA is a two-stage training paradigm that combines grounding-enhanced supervised fine-tuning (SFT) with Group Relative Policy Optimization (GRPO), thereby improving action reliability and decision consistency during closed-loop navigation.
Across three ERCP subtasks, BiliVLA achieves the best overall performance in physical phantom experiments, with a total mIoU of 0.9625, an overall action precision of 91.96\%, and an overall success rate (SR) of 84.85\%. These results indicate that integrating semantic grounding, scene-aware learning, and reward-guided optimization strengthens perception--action alignment and enables more robust autonomous biliary endoscopic navigation.
\end{abstract}

\section{Introduction}
Endoscopic intubation and cannulation in pancreatobiliary interventions, particularly in endoscopic retrograde cholangiopancreatography (ERCP), remain among the most technically demanding procedures in therapeutic endoscopy \cite{ESGE2014PEP,ercp1}. Procedural success depends on precise and stable manipulation within a narrow monocular field of view in a deformable, specular, and frequently occluded environment. Selective biliary cannulation typically requires: (i) reliable identification of the major duodenal papilla and accurate alignment of the endoscopic view or catheter tip with the papillary orifice; (ii) coordinated forward advancement combined with distal tip bending to achieve deep cannulation; and (iii) rapid re-centering and refocusing on suspected calculi or biliary targets once visual evidence becomes available. Prolonged or repeated cannulation attempts are associated with an increased risk of adverse events, most notably post-ERCP pancreatitis (PEP), and cannulation failure remains clinically significant even for experienced operators \cite{ESGE2014PEP,Cahyadi2022PEP}

Recent clinical systems have explored robotic assistance for ERCP to improve ergonomics and procedural stability \cite{Chen2026RobotERCP}. In parallel, computer vision methods have been investigated to support ERCP by detecting the ampulla and estimating cannulation difficulty directly from endoscopic images \cite{Kim2021AIERCP}. However, most autonomous endoscopic systems adopt a modular design in which perception, planning, and control are optimized independently \cite{britle1, britle2, britle3}. Learning-based endoscopic control has also been explored for tendon-driven and deformable continuum robots through deep reinforcement learning~\cite{ng2024navigation,ng2025contact,tian2025jacobian}. These methods mainly focus on geometry-driven control and do not directly unify language-conditioned semantic grounding with closed-loop action generation.
 Such decoupling can reduce robustness under scene variability and domain shifts, and it constrains the direct translation of procedural semantics into closed-loop actions.

Vision-Language-Action (VLA) models provide an alternative paradigm by integrating visual perception, language grounding, and action generation within an instruction-conditioned policy. This formulation enables semantic adaptation to operator prompts while reducing reliance on hand-engineered interfaces between modules \cite{zhong2025survey}. In general robotics, large-scale instruction-conditioned visuomotor policies demonstrate strong generalization by leveraging diverse robot datasets and internet-scale vision-language pretraining \cite{sapkota2025vision}. In surgical and endoluminal robotics, early VLA studies further demonstrate the feasibility of dual-phase fine-tuning and hierarchical language-conditioned control for long-horizon procedures \cite{EndoVLA_2025,SRT_H_2025}.


\begin{figure*}[t!]
  \centering
  \includegraphics[width=0.9 \textwidth,trim=0cm 0cm 0cm 0cm,clip]{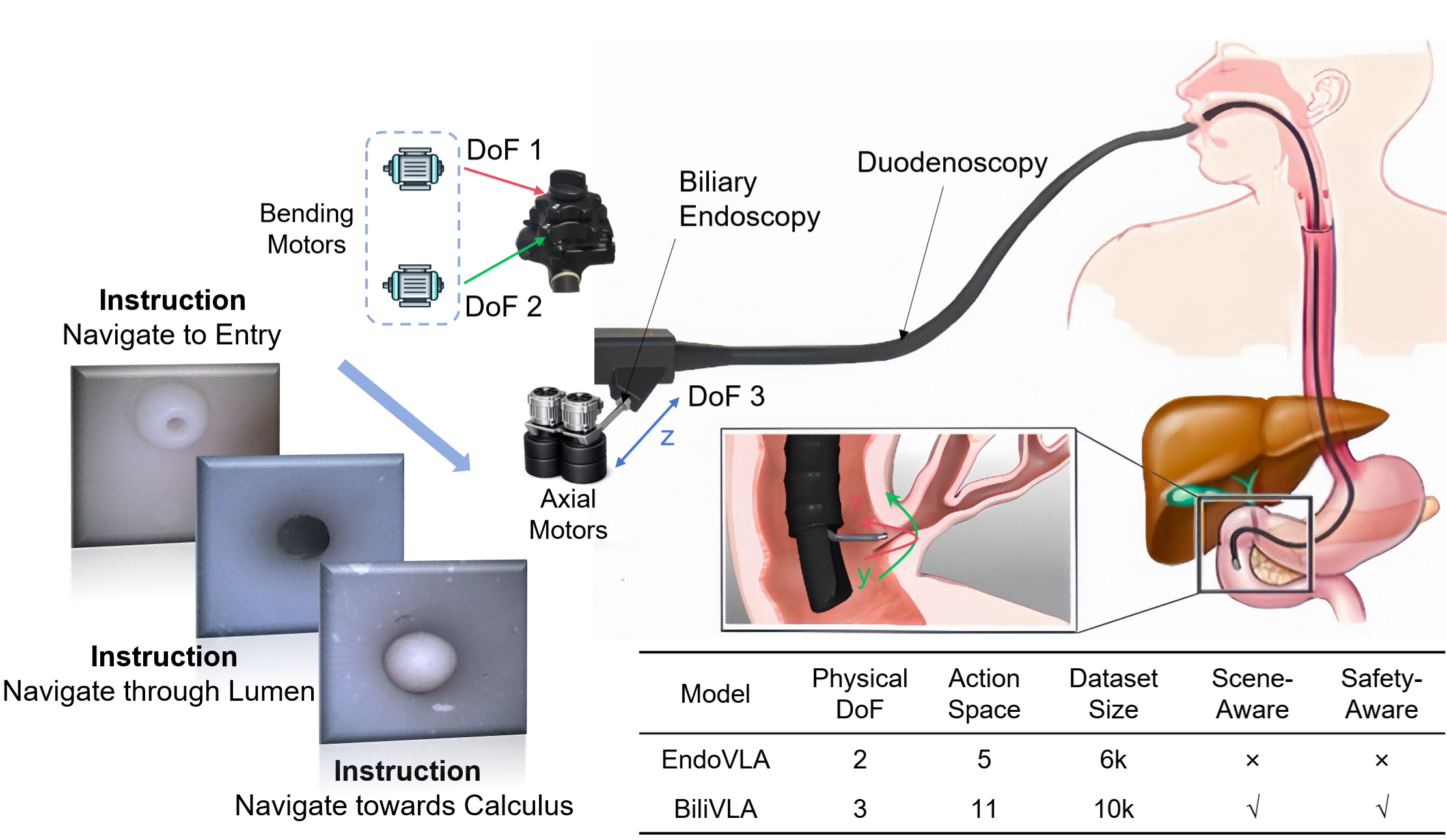}
\caption{\textbf{Instruction-guided robotic endoscopic navigation for ERCP workflows.}
Left: stage-specific language instructions with endoscopic observations.
Middle: robotic endoscope backend with bending ($x/y$) and axial ($z$) DoF.
Right: clinical ERCP navigation context with a zoomed papilla view.
Bottom-right: comparison of EndoVLA\cite{EndoVLA_2025} and BiliVLA in DoF, action space, dataset size, and scene/safety awareness.}
  \label{fig:intro_overview}
\end{figure*}


Despite substantial progress of VLA models in both general robotics and surgical applications, their direct deployment for endoscopic navigation and cannulation remains fundamentally challenging. First, considerable anatomical variability across gastrointestinal regions introduces pronounced domain shifts and substantial discrepancies in target appearance, thereby imposing stringent requirements on semantic grounding and cross-scene generalization\cite{van2025gastrointestinal}. Second, intra-procedural factors, including specular reflections, partial occlusion, and luminal wall contact, frequently cause severe visual degradation, which increases the likelihood of unintended collision and tissue injury\cite{qian2022specular}. Finally, cannulation-grade manipulation demands motion control that is precise, stable, and robust within highly constrained spatial environments\cite{ercp2}, thus requiring policy optimization that extends beyond conventional visuomotor benchmarks.

To address these challenges, we present \textbf{BiliVLA}, a scene-aware VLA framework for autonomous biliary endoscopic navigation in ERCP. We cast navigation as a goal-conditioned visuomotor decision problem that unifies semantic target understanding, visual grounding, and discrete three degrees of freedom (DoF) continuum actuation within a single policy. By explicitly conditioning action generation on scene-level semantics and introducing safety-aware supervision, the framework stabilizes target alignment and mitigates collision risks under diverse initial configurations. We adopt a two-stage training strategy that integrates grounding-enhanced supervised fine-tuning with reward-guided policy optimization, progressively improving spatial localization fidelity and action consistency. Extensive evaluations on three clinically relevant subtasks, namely entry navigation, lumen traversal, and calculus localization, demonstrating the effectiveness of the proposed framework in enabling reliable perception–action coupling for endoscopic robotic navigation. 

The core contributions of this work are threefold:

\begin{itemize}
\item We propose \textbf{BiliVLA}, a scene-aware VLA framework tailored for biliary endoscopic robots, together with a two-stage training paradigm that systematically enhances semantic reasoning, spatial grounding, and robust action generation for continuum navigation.

\item We construct the \textbf{BiliVLA-Motion} dataset, a vision--language--kinematic dataset containing 10k annotated image--motion pairs across three ERCP subtasks, providing structured supervision for multimodal endoscopic policy learning.

\item We validate the framework through physical phantom experiments, where BiliVLA achieves the best overall performance with a total mIoU of 0.9625, an overall action precision of 91.96\%, and an overall task success rate of 84.85\%, demonstrating improved perception--action alignment and more reliable task execution compared with existing baselines.

\end{itemize}

\section{Related Work}
\subsection{Robotic and AI-assisted ERCP Cannulation}
Prolonged procedures and increased PEP risk are major challenges in ERCP, with cannulation being the most critical stage, requiring high dexterity and stability to minimize complications \cite{Testoni2011DifficultCannulation,ESGE2014PEP}. Radiation exposure also poses a risk to healthcare providers, emphasizing the need for robotic automation of cannulation \cite{Cahyadi2022PEP}. Robotic systems have proven effective in facilitating cannulation, cholangiography, and biliary stent placement \cite{Chen2026RobotERCP}. In addition, artificial intelligence (AI) has been applied to detect the ampulla and predict cannulation difficulty from ERCP images, demonstrating promising accuracy and clinical relevance \cite{Kim2021AIERCP}. Despite these advances, fully automating biliary endoscopic cannulation remains challenging due to anatomical complexity, precision requirements, and the need for safe tissue interaction.

\subsection{VLA for Generalist Robot Control}
VLA policies integrate visual perception, language grounding, and action prediction through transformer policies trained on diverse robot trajectories. RT-1~\cite{RT1_2022}, RT-2~\cite{RT2_2023}, $\pi_0$~\cite{black2024pi_0}, $\pi_{0.5}$~\cite{intelligence2025pi_}, and RDT~\cite{liu2024rdt} demonstrate scalable instruction-conditioned visuomotor learning, while Open-H-Embodiment~\cite{nelson2026open} highlights the growing importance of medical-robotics datasets for foundation models. However, endoluminal deployment remains challenging because of domain shift, safety-critical visual degradation, and strict stability requirements.

\subsection{Learning-based Autonomy for Endoluminal Robotics}
Learning-based autonomy has recently advanced endoluminal and surgical robotics. Prior reinforcement-learning methods explored tendon-driven endoscope navigation, contact-aided navigation, and Jacobian exploratory control for deformable continuum robots~\cite{ng2024navigation,ng2025contact,tian2025jacobian}. In parallel, EndoVLA~\cite{ng2025endovla} demonstrates language-conditioned perception--action alignment for autonomous endoscopic tracking, while hierarchical language-conditioned imitation learning supports long-horizon step-level surgical autonomy. These studies motivate BiliVLA, which targets ERCP-oriented biliary navigation with scene-aware grounding, luminal-wall-contact recovery, and discrete 3-DoF continuum actuation.



\section{Robotic System and Dataset Collection}
\subsection{Biliary Endoscopic Robot System}
To deploy the BiliVLA model in real world, we develop a biliary endoscopic robot system based on a commercial Olympus endoscope. The endoscope is equipped with two control knobs that govern the distal tip motion in two DoF, where we actuate each knob using an independent motor, as shown in Fig.~1. To achieve insertion and retraction, we design a transmission module consisting of two grooved rollers that hold the endoscope within a shared channel. Two motors rotate the rollers in opposite directions, which generates forward and backward motion of the endoscope shaft. By coordinating the four motors, the system realizes three DoF at the endoscopic tip, including two bending DoF and one translational DoF, thus supporting autonomous navigation and target localization within the bile duct. 

The endoscopic imaging system operates at 30 frames per seconde (FPS), and all experiments are conducted with image resolution fixed at 640$\times$480 pixels. The output of BiliVLA is converted into motor angles under the assumption of linearity mapping between motor angles and distal tip bending angles, thereby driving the corresponding endoscopic tip motions. To emulate the cannulation procedure in ERCP, we employ a phantom model of the duodenum and the common bile duct as the surgical environment, and an artificial calculus is placed inside the bile duct to simulate clinical scenarios. Fig.~2 shows the hardware setup of the robotic endoscopic navigation system for ERCP procedures.

\begin{figure}[htbp]
\centering
\includegraphics[width=0.98\columnwidth,height=6cm,keepaspectratio]{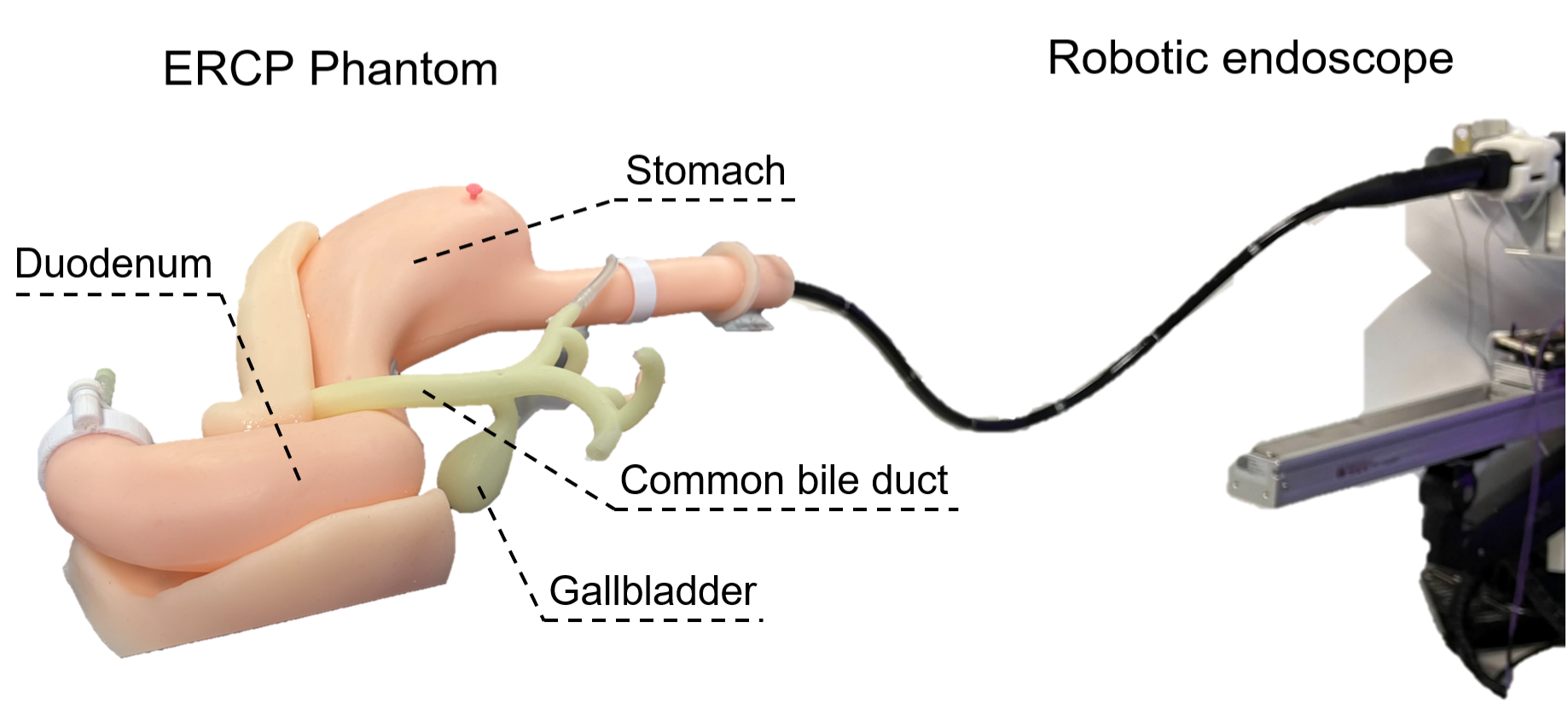}
\caption{\textbf{Hardware setup of the robotic endoscopic navigation system for ERCP procedures.} The system consists of an anatomically structured ERCP phantom and a robotic endoscope platform that enables distal tip bending and axial insertion control.}
\label{fig:hardware setup}
\end{figure}

\subsection{BiliVLA-Motion Dataset}
To address the lack of multimodal data for continuum robots, we construct the BiliVLA-Motion, a vision-language-kinematic dataset designed to train the BiliVLA model. The dataset contains 10k image-motion pairs spanning three tasks, namely entry navigation, lumen traversal, and calculus localization. Entry navigation includes 3k pairs, lumen traversal contains 2.6k pairs, and calculus localization comprises 4.4k pairs. All data are collected using the cholangioscopic robotic system in a phantom environment covering the duodenum and common bile duct.


\begin{figure}[t!]
  \centering
  \includegraphics[scale=0.17]{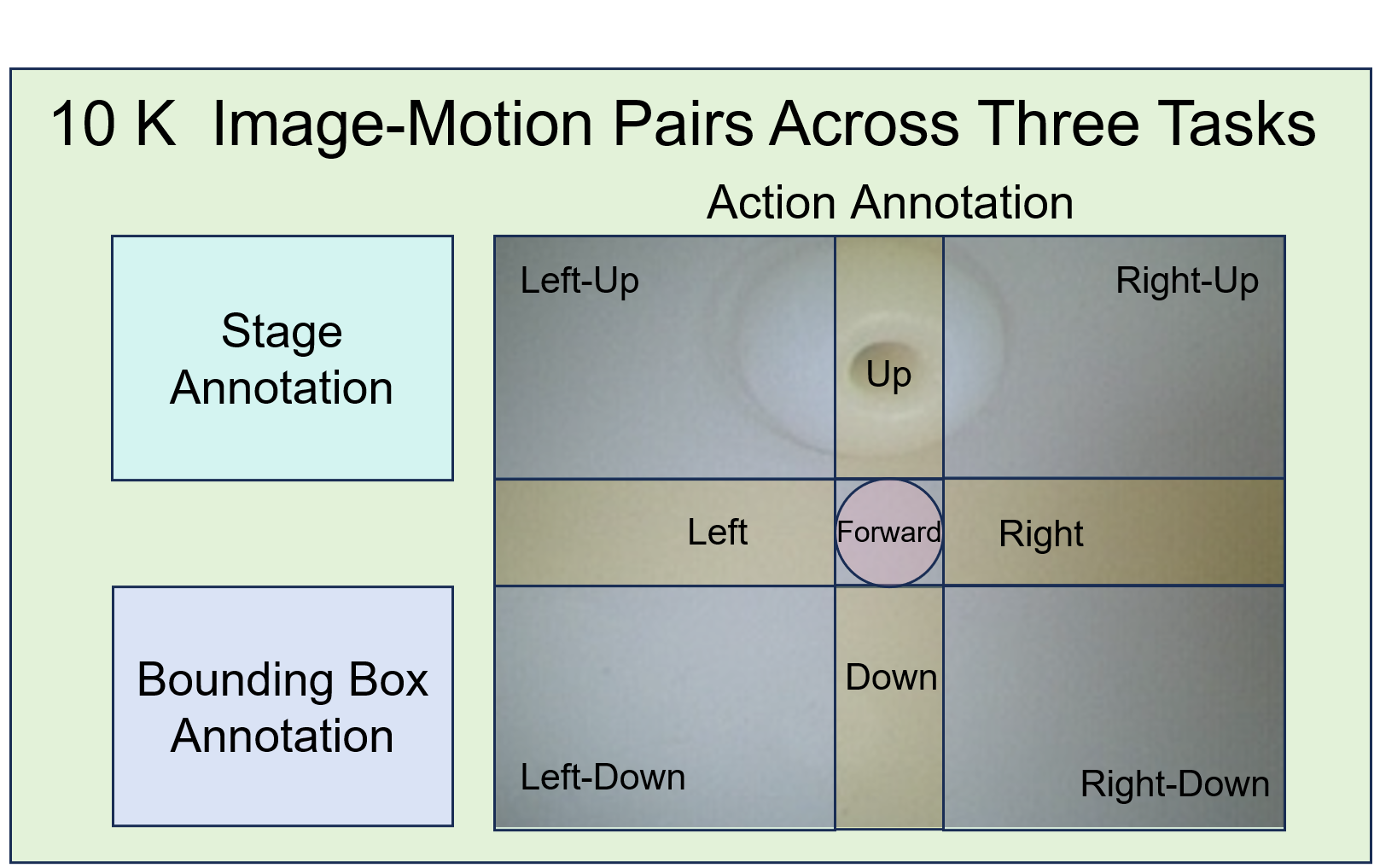}
\caption{\textbf{Generation of the BiliVLA-Motion dataset.} 
Each frame is annotated with the navigation stage, target bounding box, and a discrete motion command, forming 10K image–motion pairs across three ERCP tasks.}
  \label{Fig:1}
\end{figure}



As illustrated in Fig. 3, raw teleoperation videos are annotated with procedural stage, target bounding box, and motion command. Bounding boxes are generated by YOLOv11~\cite{khanam2024yolov11} and manually refined. Motion labels are assigned according to the target-center offset relative to a 44-pixel focus region: centered targets trigger forward motion, off-center targets trigger cardinal or diagonal bending, centered visible calculus triggers stop, and visually occluded wall-contact frames trigger backward recovery. The dataset is split into training, validation, and test sets with a ratio of 80\%, 10\%, and 10\%.

\begin{figure*}[t!]
  \centering
  \includegraphics[scale=0.28]{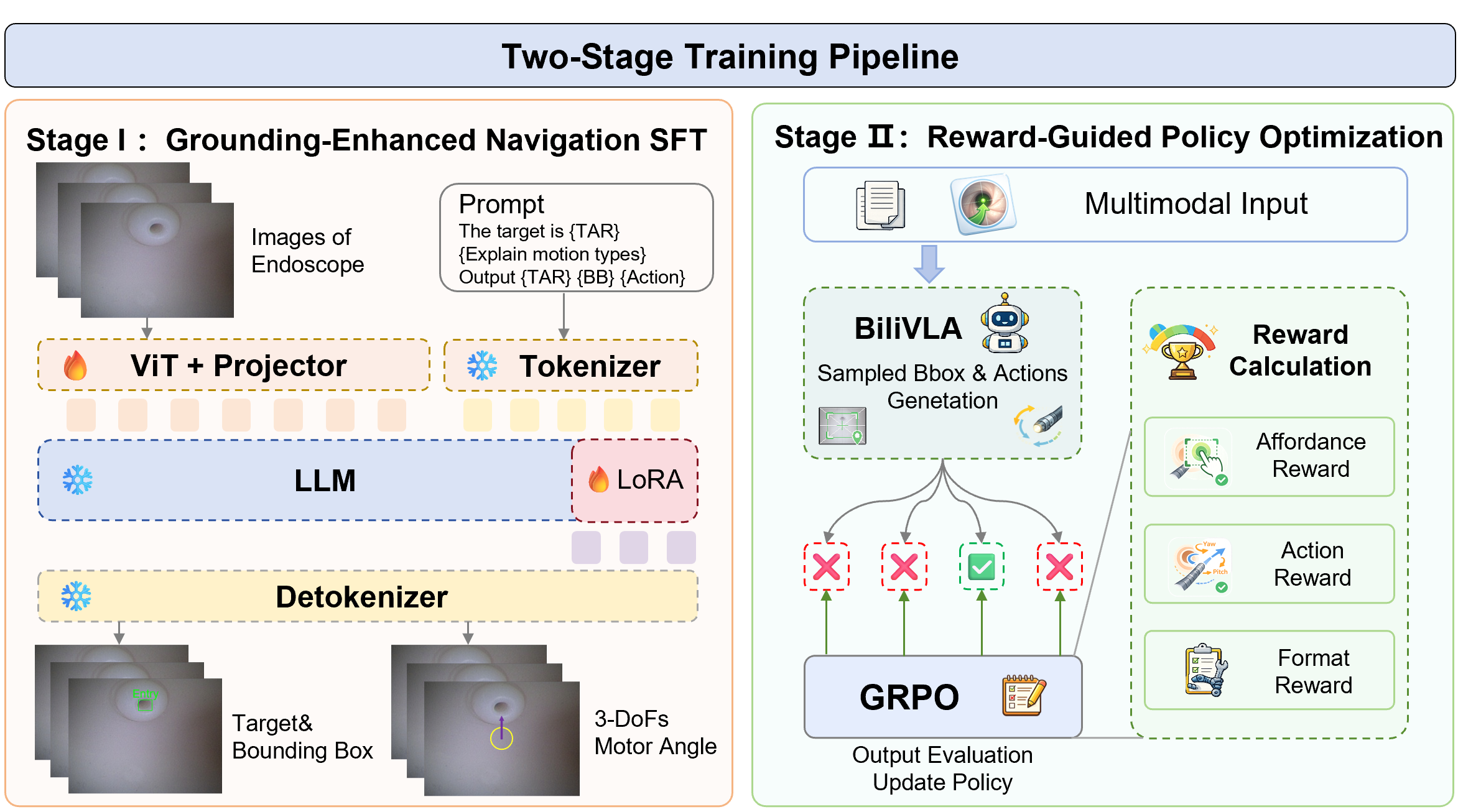}
  \caption{Overview of BiliVLA. The framework takes a monocular endoscopic observation and a stage-specific language instruction as input, and outputs the target category, grounded bounding box, and discrete 3-DoF motor command for biliary navigation. BiliVLA integrates scene-aware supervision, safety-aware recovery learning, and a two-stage training pipeline with grounding-enhanced supervised fine-tuning followed by GRPO-based reward-guided optimization.}
  \label{Fig:}
\end{figure*}

\section{Method}
\subsection{Problem Definition}

The BiliVLA framework is formulated as a target-conditioned visuomotor decision process. At each time step, the agent receives an endoscopic RGB observation and a target specification, and outputs a structured prediction that includes the target category, its spatial localization, and a control command. The objective is to guide the endoscope toward target while maintaining safe interaction with surrounding tissue. Each episode is
\begin{equation}
\mathcal{E} = \{(o_t, I, s_t)\}_{t=1}^{T},
\quad
s_t = (c_t, b_t, a_t),
\end{equation}
where $o_t \in \mathbb{R}^{H \times W \times 3}$ represents the RGB image at time step $t$, $I$ denotes the target instruction, and $s_t$ encapsulates the structured supervision signal. Specifically, $c_t$ indicates the target category (e.g., entry, lumen, or calculus), $b_t = (x_t, y_t, w_t, h_t)$ corresponds to the ground-truth bounding box in normalized image coordinates, and $a_t \in \mathcal{A}$ represents the discrete control action. The action space contains eleven motion primitives, $\mathcal{A} = $\{\text{left-up}, \text{left-down}, \text{right-up}, \text{right-down}, 
\text{left}, \text{right}, \text{up}, \text{down}, 
\text{forward}, \text{backward}, \text{stop}\}, which jointly parameterize planar bending, axial translation, and motion termination, enabling structured actuation under instruction-conditioned control.

Each action corresponds to an incremental motor command in a three-angle actuation space. Specifically, we represent low-level actuation by
\begin{equation}
\Delta\boldsymbol{\theta}_t = (\Delta \theta_{x,t}, \Delta \theta_{y,t}, \Delta \theta_{z,t}) \in \mathbb{R}^{3},
\end{equation}
where $\Delta \theta_{x,t}$ and $\Delta \theta_{y,t}$ parameterize bending in two orthogonal directions of the distal tip, and $\Delta \theta_{z,t}$ parameterizes insertion or retraction. Each discrete action $a_t$ deterministically maps to a fixed angular increment $\Delta\boldsymbol{\theta}(a_t)$, and the corresponding motor update follows $\boldsymbol{\theta}_{t+1} = \boldsymbol{\theta}_t + \Delta\boldsymbol{\theta}(a_t)$.

The control objective is to iteratively reduce the spatial discrepancy between the detected target and the image center. The corresponding target center is computed as $p_t = (x_t + w_t/2, \, y_t + h_t/2)$, while the image center is denoted by $p_c$. At each time step, the selected action is expected to decrease the Euclidean distance $\|p_t - p_c\|_2$ in the image plane, subject to predefined termination conditions. The resulting optimal decision rule at time step $t$ is defined as
\begin{equation}
a_t^*=
\begin{cases}
\text{stop}, & \text{calculus is centered},\\
\arg\min_{a\in\mathcal{A}}\|p_t(a)-p_c\|_2, & \|p_t-p_c\|_2>\tau,\\
\text{forward}, & \|p_t - p_c\|_2 \le \tau. \\
\end{cases}
\end{equation}
where $p_t(a)$ denotes the estimated target center after applying action $a$ for one time step. This formulation integrates visual grounding, spatial alignment, and discrete endoscopic actuation into a unified policy learning framework.

\subsection{Scene-aware Policy Learning}
To improve scene understanding and target consistency, a scene-aware supervision mechanism is incorporated into the training process. Unlike purely geometric alignment, this strategy explicitly conditions the policy on semantic target information and enforces target-aware prediction in the output space.

At each time step, the policy is conditioned on a task-level textual instruction to enhance scene awareness. Let $I$ denote the instruction associated with a given task. 
We model $I$ as a task-conditioned prompt generated by a template function
\begin{equation}
I = \mathcal{T}(c^{\text{task}}),
\end{equation}
where $c^{\text{task}} \in \mathcal{C}$ specifies the target category, and $\mathcal{T}(\cdot)$ produces a structured textual prompt that encodes both the target identity and the required output schema, including normalized bounding box coordinates and discrete action components. The policy $\pi_\theta$ parameterized by $\theta$ jointly predicts the target category, bounding box, and action,
\begin{equation}
(\hat{c}_t, \hat{b}_t, \hat{a}_t) = \pi_\theta(o_t, I),
\end{equation}
where the predictions $(\hat{c}_t, \hat{b}_t, \hat{a}_t)$ are obtained via autoregressive conditional decoding.

By explicitly conditioning on target semantics in both input and output spaces, the policy learns to align visual perception with task-relevant scene context. This scene-aware formulation reduces target ambiguity and improves robustness across heterogeneous anatomical environments.

\subsection{Safety-aware Policy Learning}
To promote safe behavior during autonomous navigation, we incorporate safety-aware supervision into the training process. In ERCP procedures, excessive proximity to surrounding tissue can induce occlusion or pronounced visual degradation in the endoscopic view. We collectively refer to these unsafe visual patterns as luminal wall contact.

For frames detected as luminal wall contact, we assign a conservative recovery label and use it as the supervisory signal during training. Specifically, the bounding box is set to cover the entire image, \mbox{$b_t^{\text{s}} = (0,0,1,1)$}, while the control command is defined as \mbox{$a_t^{\text{s}} = \text{backward}$}. For all remaining frames, we apply the nominal supervision derived from geometric alignment, denoted by $(\hat{b}_t, \hat{a}_t)$. The resulting training target is therefore defined as
\begin{equation}
(b_t, a_t) =
\begin{cases}
(b_t^{\text{s}}, a_t^{\text{s}}), & o_t \text{ indicates luminal wall contact}, \\
(\hat{b}_t, \hat{a}_t), & \text{otherwise}.
\end{cases}
\end{equation}

This supervision scheme injects an explicit safety prior into policy learning. By exposing the model to retreat labels under luminal wall contact, the resulting policy learns to recover from unsafe proximity and maintain a stable and collision-averse navigational configuration.

\subsection{Two-Stage Training Framework for VLA}
To address the strong coupling among semantic target understanding, spatial localization, and control decision-making, we design a two-stage training framework to systematically enhance cross-task generalization in endoscopic navigation, as illustrated in Fig.~3. The framework adopts a hierarchical architecture that integrates scene-level semantic supervision, spatial grounding data, and high-quality biliary endoscopic navigation demonstrations. By combining SFT and RFT, the framework progressively improves scene understanding, spatial reasoning, and motion planning capabilities in complex endoscopic environments, ultimately enabling a VLA-based endoscopic navigation agent.

\textbf{Stage I: Grounding-Enhanced Navigation SFT.} In the first stage, we employ Low-Rank Adaptation (LoRA)\cite{hu2022lora} to efficiently fine-tune the large language model (LLM). The backbone LLM and tokenizer remain frozen, thereby preserving pretrained multimodal priors. Endoscopic images are encoded by a trainable Vision Transformer (ViT) together with a multilayer perceptron (MLP) projector, which maps visual features into the language embedding space of the LLM. The textual instruction is processed by the tokenizer of the LLM. A detokenizer transforms the predicted tokens into executable representations, producing both visual grounding results for target localization and three DoF motor-angle commands for navigation control. The training dataset comprises multi-view images, spatial grounding annotations, scene-level semantic information, and motion commands. This design strengthens foundational visual understanding, spatial relation modeling, and action generation capabilities. These components establish robust navigation-related representations and provide a stable semantic and instruction-level backbone for subsequent reinforcement learning.

\textbf{Stage II: Reward-Guided Policy Optimization.} In the second stage, we perform reward-guided RFT to further improve prediction accuracy, decision reliability, and motion consistency. For each input observation, the model samples multiple candidate outputs consisting of target category predictions, normalized bounding box estimates, and discrete three-DoF action commands. These sampled trajectories are evaluated using a structured reward function that captures both perceptual alignment and action feasibility. The policy is subsequently optimized with Group Relative Policy Optimization (GRPO) \cite{guo2025deepseek}, which updates the model by comparing the relative rewards of sampled candidates within each group.

Formally, for each input $\mathbf{u}$, a group of $K$ samples $\{\mathbf{s}_i\}_{i=1}^{K}$ is drawn from $\pi_\theta$. Each sample is assigned a scalar reward $r_i = R(\mathbf{u}, \mathbf{s}_i)$. To avoid reliance on an external value function, GRPO estimates normalized advantages within each group by contrasting individual rewards against the group-level statistics, thereby providing a stable and variance-reduced learning signal for policy optimization.

The overall reward function is defined as
\begin{equation}
R(\mathbf{u}, \mathbf{s})
=
R_{\text{bbox}}
+
R_{\text{act}}
+
 R_{\text{fmt}}.
\end{equation}

The reward of the bounding box is defined using the Intersection-over-Union (IoU) between the predicted bounding box $B_{\text{pred}}$ and the ground-truth box $B_{\text{gt}}$:
\begin{equation}
R_{\text{bbox}}
=
\operatorname{IoU}(B_{\text{pred}}, B_{\text{gt}})
=
\frac{|B_{\text{pred}} \cap B_{\text{gt}}|}
{|B_{\text{pred}} \cup B_{\text{gt}}|}.
\end{equation}
This term provides dense spatial feedback to improve localization accuracy.

The action reward evaluates the correctness of the discrete 3-DoF command:
\begin{equation}
R_{\text{act}}
=
\begin{cases}
1, & \text{if } a_{\text{pred}} = a_{\text{gt}}, \\
0, & \text{otherwise}.
\end{cases}
\end{equation}
This binary signal enforces strict alignment between predicted and reference actions.

The format reward ensures structural validity of the generated output:
\begin{equation}
R_{\text{fmt}}
=
\begin{cases}
1, & \text{if output follows predefined schema}, \\
0, & \text{otherwise}.
\end{cases}
\end{equation}

\begin{figure*}[t!]
  \centering
  \includegraphics[scale=0.24]{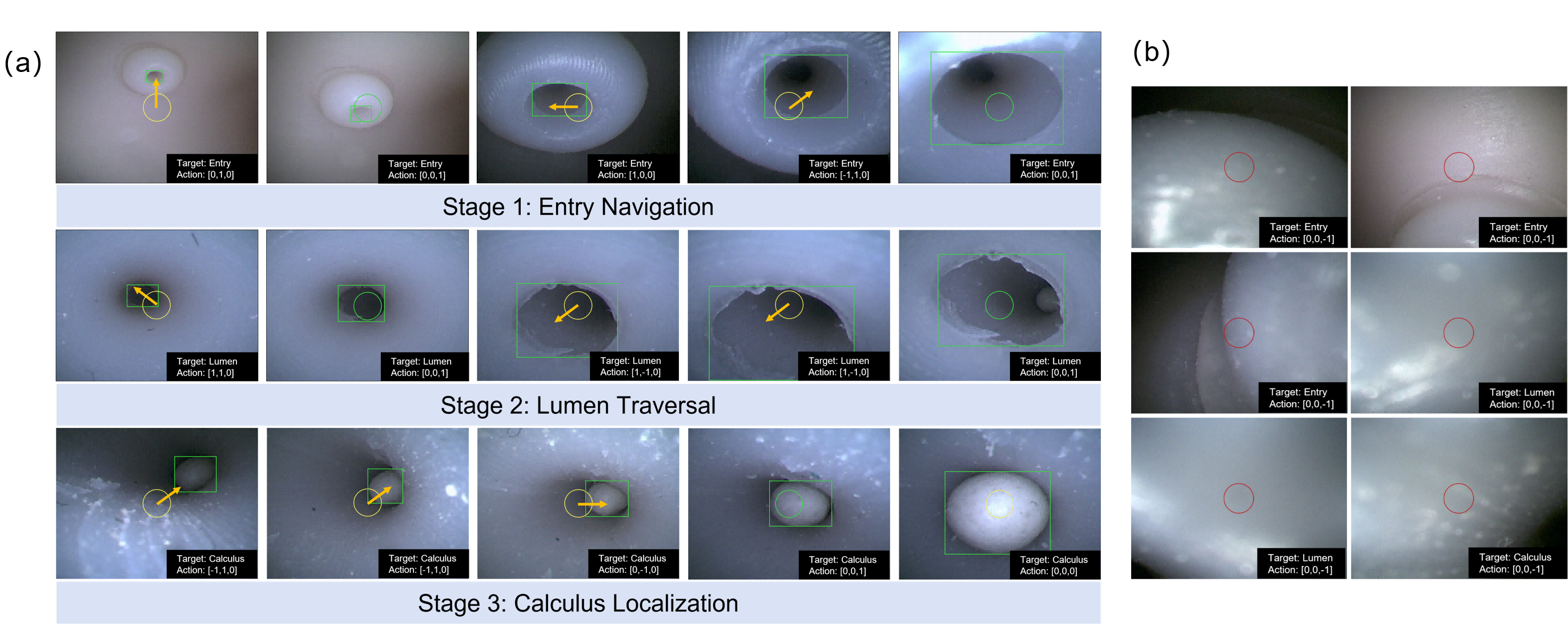}
\caption{\textbf{Qualitative results of BiliVLA in ERCP navigation.} 
(a) Representative predictions across three navigation stages: entry navigation, lumen traversal, and calculus localization. Green boxes denote grounded targets, yellow arrows indicate the predicted bending direction, and the circle marker represents axial motion (insertion/stop/retraction). 
(b) Safety-aware recovery under luminal wall contact, where the policy generates backward actions to avoid boundary collisions.}
  \label{Fig:}
\end{figure*}

This constraint guarantees that the generated prediction remains syntactically parsable and directly executable by the control system.

Through GRPO-based optimization, the policy is encouraged to jointly satisfy spatial grounding accuracy, action correctness, and structural consistency, thereby improving robustness in complex endoscopic environments.

\section{Experiment}

\subsection{Experiment Setup}
To evaluate BiliVLA under diverse initial conditions, the biliary endoscope is inserted through the working channel of a duodenoscope and initialized at a random pose in the vicinity of the papilla phantom. The experimental protocol follows the standard ERCP procedure and is decomposed into three sequential phases with distinct objectives. In the entry navigation phase, the agent is required to align the endoscopic view with the papilla and advance toward the papillary orifice to establish ductal access. During lumen traversal, the endoscope progresses along the ductal lumen while maintaining stable and centered navigation. In the calculus localization phase, the system aligns the visual axis with the target calculus and halts in close proximity, thereby preparing for subsequent interventional manipulation.

\subsection{Implementation Details}
We adopt a two-stage training pipeline consisting of SFT followed by RFT. Specifically, BiliVLA is first fine-tuned on the BiliVLA-Motion dataset using Unsloth\cite{Unsloth2024}, with Qwen3-VL-8B\cite{bai2025qwen3} as the backbone and LoRA for parameter-efficient adaptation. Training is conducted on four NVIDIA RTX A6000 GPUs. During SFT, we set the sequence length to 1024 and use an effective batch size of 8 via gradient accumulation. Optimization is performed with 8-bit AdamW and a linear learning-rate scheduler for 2k steps, with periodic evaluation and checkpointing.

We then perform RFT using TRL GRPO\cite{vonWerra2024trl}, initialized from the SFT-adapted LoRA weights while keeping the vision encoder frozen. The reward function jointly enforces output format validity, bounding-box affordance, and action consistency. RFT relies on sampled generations to estimate reward signals. All inference experiments are conducted on a single NVIDIA RTX 5090 GPU. During closed-loop deployment, BiliVLA performs inference at 2~Hz, enabling real-time action prediction for physical endoscopic navigation.
\subsection{Experiment Results}

\begin{table*}[htbp]
\centering
\caption{mIoU of bounding box prediction,  PR (\%) of action prediction, and SR (\%) across different tasks. }
\resizebox{\textwidth}{!}{
\begin{tabular}{l|ccc|ccc|ccc|ccc}
\toprule
\multirow{2}{*}{Model} &
\multicolumn{3}{c|}{Entry Navigation} &
\multicolumn{3}{c|}{Lumen Traversal} &
\multicolumn{3}{c|}{Calculus Localization} &
\multicolumn{3}{c}{Total} \\
\cmidrule(lr){2-4}
\cmidrule(lr){5-7}
\cmidrule(lr){8-10}
\cmidrule(lr){11-13}
& mIoU & PR(\%) & SR(\%) 
& mIoU & PR(\%) & SR(\%) 
& mIoU & PR(\%) & SR(\%) 
& mIoU & PR(\%) & SR(\%) \\
\midrule
Imitation Learning & 0.7315      & 74.17 & 27.27 & 0.7151      & 76.92 & 54.55 & 0.7894      & 78.80 & 45.45 & 0.7657      & 77.31 & 42.42 \\
Qwen3-VL           & 0.7595 & 81.77  & 43.28     & 0.7352 & 81.92  & 64.83     & 0.8465 & 80.60 & 53.42     & 0.8117 & 81.14 & 51.82     \\
EndoVLA            & 0.8189 & 84.49 & 51,69     & 0.8124 & 84.62 & 71.29     & 0.8537 & 84.26 & 59.93     & 0.8406 & 84.44 & 58.86     \\
BiliVLA (w/o GRPO) & 0.8832 & 86.57 & 54.55 & 0.9480 & 87.50 & 72.73 & 0.9782 & 90.33 & 63.64 & 0.9488 & 89.00 & 63.64 \\
BiliVLA (w/o Scene Aware)
                   & 0.8980 & 88.69 & 63.64 & 0.9477 & 89.62 & 81.82 & 0.9798 & 91.44 & 72.73 & 0.9430 & 90.49 & 72.73 \\
Ours               & \textbf{0.9162} & \textbf{90.82} & \textbf{72.73} 
                   & \textbf{0.9630} & \textbf{91.46} & \textbf{100.00} 
                   & \textbf{0.9816} & \textbf{92.55} & \textbf{81.82} 
                   & \textbf{0.9625} & \textbf{91.96} & \textbf{84.85} \\
\bottomrule
\end{tabular}
}
\end{table*}

We conducted a systematic evaluation on three representative tasks, namely entry navigation, lumen traversal, and calculus localization, to examine the model’s unified perception and decision-making capability. Transitions between tasks are governed by stage-specific language instructions, enabling seamless switching within a single instruction-conditioned policy. 

Fig.~5(a) presents qualitative results for all tasks. The circular marker encodes the translational DoF along the forward--backward axis, where green denotes forward motion, yellow indicates stop, and red represents backward motion. The arrow marker corresponds to the two bending DoF, and its orientation specifies the bending direction. Target instances are visualized using bounding boxes that convey both spatial location and semantic category. This structured representation strengthens spatial reasoning and improves object recognition and localization, which in turn enhances navigation performance.

Fig.~5(b) illustrates the behavioral impact of safety-aware supervision under luminal wall contact conditions. When the endoscope approaches the luminal boundary and the visual observation presents unsafe proximity patterns, such as dominant near-field wall regions or severe occlusion caused by physical contact, the policy shifts to a conservative recovery mode acquired during training. In accordance with the supervision strategy described in Sec.~IV-B, the predicted bounding box expands to cover the entire image, and the control command switches to backward motion. This response embodies the injected safety prior: rather than persisting in forward exploration under high-risk or ambiguous visual states, the agent deliberately retreats to restore a stable and safer viewing configuration. Qualitative results indicate that the policy preserves semantic target awareness while mitigating sustained luminal wall contact, demonstrating that safety-aware policy learning effectively yields collision-averse behavior during real-world interaction.

Performance is evaluated using three metrics: mean Intersection-over-Union (mIoU) for bounding box prediction, action precision rate (PR), and task success rate (SR), which together reflect spatial localization accuracy and control effectiveness. The mIoU and PR metrics are computed on a held-out test set, while SR is measured through real-world deployment, where each task is executed 11 times with a maximum horizon of 50 steps per trial. Each task follows a task-specific success criterion: entry navigation succeeds when the endoscope reaches the duodenal papilla entry; lumen traversal succeeds upon discovering the target calculus; and calculus localization requires the calculus to appear near the image center with clear visibility.

Table~1 reports the quantitative results across all tasks.  The imitation learning (IL) baseline obtains a total mIoU of 0.7657 and an overall action precision of 77.31\%, but its overall SR is only 42.42\%. This suggests that direct supervised imitation provides limited robustness in both visual grounding and action execution under complex endoscopic navigation conditions. Qwen3-VL and EndoVLA achieve overall SRs of 51.82\% and 58.86\%, respectively, indicating that general VLA-based policies can produce executable actions in part of the trials. Nevertheless, their lower mIoU scores and moderate SRs reveal limited perception--action coupling under complex endoscopic navigation conditions.

Our approach delivers the strongest overall task performance, particularly in terms of action reliability and real-world completion rates. It achieves the highest total mIoU of 0.9625, the highest overall action precision of 91.96\%, and the highest overall SR of 84.85\%. For entry navigation, lumen traversal, and calculus localization, our approach achieves action precision values of 90.82\%, 91.46\%, and 92.55\%, respectively. The corresponding SRs are 72.73\%, 100.00\%, and 81.82\%, consistently outperforming all competing methods across different task scenarios. These results demonstrate that the proposed scene-aware supervision and reinforcement-enhanced optimization promote stronger perception--action alignment, leading to more stable and effective closed-loop behavior in real-world deployments.

\subsection{Ablation Study}

The last three rows of Table 1 present the ablation results, which analyze the contributions of GRPO optimization and the scene-aware module. Removing GRPO, denoted as BiliVLA (w/o GRPO), preserves relatively strong localization performance, with mIoU scores of 0.8832, 0.948, and 0.9782 across the three tasks. Nevertheless, both PR and SR decrease consistently compared to the full model. In particular, the overall SR drops from 84.85\% to 63.64\%, underscoring the role of GRPO in improving action reliability and stabilizing real-world execution.

A similar pattern is observed when the scene-aware component is removed, denoted as BiliVLA (w/o Scene Aware). Although mIoU and PR remain competitive, the SRs decline to 63.64\%, 81.82\%, and 72.73\% on the three tasks, resulting in an overall SR of 72.73\%. This degradation suggests that scene-aware modeling  promotes consistent decision-making under diverse spatial layouts.

In contrast, the complete model achieves the best performance across all metrics, obtaining the highest overall mIoU of 0.9625, PR of 91.96\%, and SR of 84.85\%. These findings indicate that GRPO optimization and scene-aware perception provide complementary advantages and jointly enable robust and reliable task execution.

\section{Conclusion}
We presented BiliVLA, a scene-aware VLA framework for ERCP-oriented biliary endoscopic navigation. BiliVLA unifies stage-specific semantic objectives, visual grounding, and discrete 3-DoF continuum actuation, while incorporating GRPO-based optimization and luminal-wall-contact recovery supervision. Experiments on three phantom ERCP subtasks demonstrate that BiliVLA consistently improves visual grounding, action prediction, and real-world task completion over supervised and VLA baselines, achieving a total mIoU of 0.9625, 91.96\% PR, and 84.85\% SR.  Future work will extend evaluation to ex vivo tissue and further improve safety and robustness under more diverse anatomical variations and visual degradations.

\balance
\bibliographystyle{IEEEtran}

\bibliography{bibliography}

\end{document}